\pdfoutput=1
\documentclass[11pt]{article}

\usepackage[final]{acl}

\usepackage{times}
\usepackage{latexsym}

\usepackage{booktabs}
\usepackage{hyperref}
\usepackage{url}
\usepackage{siunitx}
\usepackage{multirow}
\usepackage{amssymb}   % For \square symbol

\usepackage[T1]{fontenc}
\usepackage{placeins}
\usepackage[utf8]{inputenc}
\usepackage{microtype}
\usepackage{afterpage}
\usepackage{inconsolata}
\usepackage{graphicx}

\title{How Real Are Synthetic Therapy Conversations? \\Evaluating Fidelity in Prolonged Exposure Dialogues}

\author{
  Suhas BN\textsuperscript{1} \quad
  Dominik Mattioli\textsuperscript{1} \quad
  Andrew M. Sherrill\textsuperscript{2} \\
  Rosa I. Arriaga\textsuperscript{3} \quad
  Chris W. Wiese\textsuperscript{4} \quad
  Saeed Abdullah\textsuperscript{1}
   \\
  \textsuperscript{1}College of Information Sciences and Technology, Penn State University, USA \\
  \textsuperscript{2}Department of Psychiatry and Behavioral Sciences, Emory University, USA \\
  \textsuperscript{3}School of Interactive Computing, Georgia Tech, USA \\
  \textsuperscript{4}School of Psychology, Georgia Tech, USA \\
  \texttt{\{bnsuhas,saeed\}@psu.edu},  \texttt{andrew.m.sherrill@emory.edu}
}

\begin{document}
\maketitle
\begin{abstract}
Synthetic data adoption in healthcare is driven by privacy concerns, data access limitations, and high annotation costs. We explore synthetic Prolonged Exposure (PE) therapy conversations for PTSD as a scalable alternative for training clinical models. We systematically compare real and synthetic dialogues using linguistic, structural, and protocol-specific metrics like turn-taking and treatment fidelity. We introduce and evaluate PE-specific metrics, offering a novel framework for assessing clinical fidelity beyond surface fluency. Our findings show that while synthetic data successfully mitigates data scarcity and protects privacy, capturing the most subtle therapeutic dynamics remains a complex challenge. Synthetic dialogues successfully replicate key linguistic features of real conversations, for instance, achieving a similar Readability Score (89.2 vs. 88.1), while showing differences in some key fidelity markers like distress monitoring. This comparison highlights the need for fidelity-aware metrics that go beyond surface fluency to identify clinically significant nuances. Our model-agnostic framework is a critical tool for developers and clinicians to benchmark generative model fidelity before deployment in sensitive applications. Our findings help clarify where synthetic data can effectively complement real-world datasets, while also identifying areas for future refinement.
\end{abstract}

\begin{figure*}
    \centering
    \includegraphics[width=\textwidth, trim={0, 0, 0, 250}, clip]{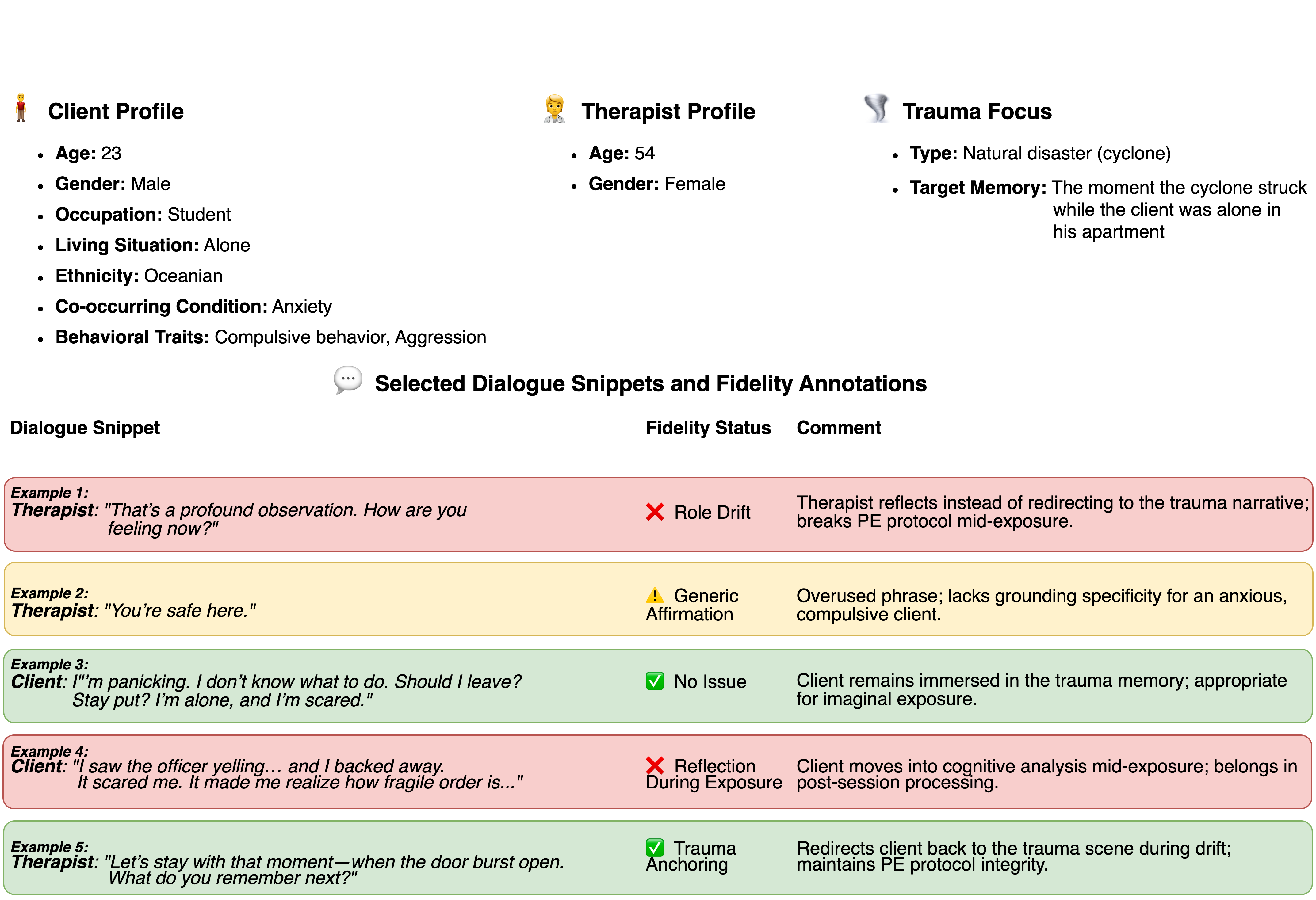}
    \caption{Selected annotated examples from synthetic PE therapy sessions. Each dialogue segment is evaluated for fidelity adherence based on PE protocol guidelines. Despite structural fluency, subtle violations like role drift (Ex. 1) and premature reflection (Ex. 4) highlight limitations in fidelity that escape automated scoring.}
    \label{fig:fidelity_example}
\end{figure*}

\section{Introduction}
\vspace{-0.1cm}
Training machine learning models in sensitive domains like healthcare remains a challenge \citep{giuffre2023harnessing, kokosi2022synthetic}. Access to real clinical conversations, which are crucial for modeling tasks like mental health diagnostics and therapeutic dialogue understanding, is severely limited by high annotation costs, patient privacy concerns \citep{kokosi2022synthetic, bn2022privacy, bn23_interspeech}, and ethical constraints on data sharing. Synthetic data has emerged as a promising alternative, offering scalability and privacy preservation while reducing dependence on real-world annotations \citep{aher2023using}. 

In trauma-focused mental health care, particularly Prolonged Exposure (PE) therapy for PTSD, large language models (LLMs) can generate synthetic therapy dialogues at scale. However, questions remain about whether these dialogues capture more than surface-level fluency, specifically the subtle dynamics of therapeutic fidelity such as emotional pacing, avoidance management, and protocol adherence \citep{4shen_2024}. To address this, we develop and validate methods to measure clinically relevant fidelity in generated dialogues, moving beyond standard metrics like coherence or perplexity. Prior work shows that while synthetic PE sessions can convincingly mimic real sessions in tone and structure \citep{bn2025thousand}, they may still commit fidelity lapses, such as premature reflection or reinforcement of avoidance, which often go unnoticed by both automatic metrics and non-clinical annotators \citep{1chiu2024, 5zhang_2024, lee2024counselingeval}. 

For instance, a therapist saying \textit{``That’s a really powerful insight''} mid-exposure may appear empathic but violates PE protocol by derailing trauma processing (see Fig. \ref{fig:fidelity_example}). Without clinical expertise, both humans and automated metrics tend to overestimate fidelity. 

We introduce a fidelity-focused lens that evaluates dialogues on multiple dimensions: linguistic coherence, adherence to PE protocols, and the therapist's navigation of key clinical interactions (e.g., managing avoidance, SUDS (Subjective
Units of Distress) monitoring). This framework integrates automated scoring and expert clinical assessment for a more rigorous evaluation of synthetic dialogue quality. 

We present the first large-scale, multi-dimensional comparison of real and synthetic PE therapy sessions, analyzing their linguistic, structural, and clinical fidelity characteristics. Our findings demonstrate the current strengths and limitations of synthetic dialogues and inform future improvements in generation and evaluation methods. Our study contributes to bridging the real-synthetic data gap and provides a roadmap for advancing synthetic dialogue in sensitive healthcare domains.

\section{Motivation}
\vspace{-0.1cm}

Developing synthetic clinical dialogues is essential to overcome persistent barriers in mental health AI. In the context of Prolonged Exposure (PE) therapy, four challenges stand out:  
(1) \textbf{data scarcity}, due to the high cost and effort of annotation;  
(2) \textbf{privacy constraints}, which limit access to sensitive patient narratives;  
(3) \textbf{lack of diversity}, with datasets often failing to capture varied trauma types and demographics; and  
(4) \textbf{evaluation inconsistency}, as fidelity assessment lacks standardized benchmarks.  

This work directly addresses these issues by generating and evaluating synthetic PE dialogues that preserve protocol fidelity, enable scalable annotation, and support more equitable and robust training data for mental health applications.

\section{Related Work}
\vspace{-0.1cm}
While synthetic clinical conversation datasets have advanced evaluations for general counseling and CBT, work specific to Prolonged Exposure (PE) and structured trauma-focused therapy is limited. Most studies focus on (1) synthetic dataset generation, (2) evaluation metrics beyond lexical similarity, and (3) human-in-the-loop validation, but generally lack PE-specific considerations.

\subsection{Synthetic Dataset Generation and Evaluation}
\vspace{-0.1cm}
Recent work has explored synthetic clinical dialogue generation and assessment. BOLT evaluates LLM-generated therapist behaviors in general counseling \citep{1chiu2024}, while SimPsyDial benchmarks synthetic data using the Working Alliance Inventory \citep{2qiu2024}. CPsyCoun reconstructs dialogues for evaluation with BERTScore, GPTScore, and qualitative review \citep{5zhang_2024, zhang2019bertscore}. Other studies focus on counselor style \citep{6xie_2024} and data augmentation \citep{7k_2024}. While The Pursuit of Empathy highlights the importance of empathy-aware evaluation \citep{suhas2025pursuit}, Thousand Voices of Trauma introduces structural variation \citep{bn2025thousand}, but most evaluations compare to general counseling or CBT \citep{3lee_2024, 4shen_2024} and rarely assess trauma-specific markers or PE’s structural fidelity, such as avoidance handling and imaginal exposure sequences. Our work directly addresses this by evaluating whether synthetic PE sessions capture clinical realism per PE protocols (e.g., SUDS, avoidance redirection), highlighting procedural errors missed by standard metrics.

\subsection{Beyond Lexical Metrics}
\vspace{-0.1cm}
Evaluation is shifting from lexical metrics like BLEU \citep{papineni2002bleu} to semantic ones such as BERTScore and GPTScore \citep{5zhang_2024, 6xie_2024}. Psychological measures (CTRS, PANAS) in COUNSELINGEVAL assess empathy and coherence \citep{3lee_2024}, and interactional features like PQA are explored \citep{4shen_2024}. However, most metrics are not adapted for trauma therapy, and methods like Dynamic Time Warping (DTW), which are potentially useful for structured PE flows, remain underused. While some qualitative reviews consider emotional tone and goal alignment \citep{5zhang_2024, 6xie_2024}, and structured questioning is explored \citep{9ren_2024}, robust frameworks for quantitative evaluation of PE-specific structure are lacking.

\subsection{Human-in-the-Loop Validation and Engagement Metrics}
\vspace{-0.1cm}
Clinician validation is now common, as in COUNSELINGEVAL \citep{3lee_2024}, PsyDT \citep{6xie_2024}, and CPsyCoun \citep{5zhang_2024}, which assess quality, empathy, and safety. Metrics for engagement and personalization are also used \citep{3lee_2024,5zhang_2024,6xie_2024}. Prior work covers empathy \citep{8morris_2018} and structured workflows \citep{9ren_2024}. However, human evaluation remains focused on general counseling or CBT, rarely addressing PE-specific components like trauma cue processing or protocol adherence compared to real PE data.

\subsection{Bridging the Gap in PE Therapy Evaluation}
\vspace{-0.1cm}
PE is a first-line treatment for PTSD \citep{sherrill2019, varkovitzky2018effectiveness, rauch2021intensive, ragsdale2020advances, evans2020understanding, yasinski2017intensive}. Despite progress in synthetic dialogue generation \citep{2qiu2024, 5zhang_2024, 6xie_2024}, evaluation metrics \citep{3lee_2024, 5zhang_2024, 6xie_2024}, and human validation, a gap remains in fidelity assessment for trauma-focused therapies like PE \citep{1chiu2024, 2qiu2024, 3lee_2024, 4shen_2024, 5zhang_2024, 6xie_2024, 10hu_2024}. Current approaches lack metrics and focus tailored to PE’s unique conversational structure, including avoidance handling, trauma cue processing, and imaginal exposure flow, and they do not address challenges such as patient emotional dysregulation \citep{2qiu2024, 3lee_2024, 4shen_2024, 5zhang_2024, 6xie_2024, 7k_2024}. Our work systematically evaluates synthetic data for alignment with PE structure and clinical validity, identifying areas for improving synthetic PE dialogue quality (see Table~\ref{tab:therapist_evaluation}, Appendix~\ref{sec:appendix-metric-definitions}).

\begin{table*}[ht]
    \centering
    \vspace{0.2cm}
    \footnotesize{
    \begin{tabular}{l c c c}
        \toprule
        \textbf{Metric} & \textbf{Yes} & \textbf{No} & \textbf{N/A} \\
        \midrule
        Therapist explained rationale for imaginal? & $\square$ & $\square$ & $\square$ \\
        Therapist gave client instructions to carry out imaginal? & $\square$ & $\square$ & $\square$ \\
        Hotspots procedure and rationale introduced? & $\square$ & $\square$ & $\square$ \\
        Therapist helped patient to identify hotspots? & $\square$ & $\square$ & $\square$ \\
        Therapist oriented the client to imaginal planned for that session? & $\square$ & $\square$ & $\square$ \\
        Therapist monitored SUDS ratings about every 5 minutes? & $\square$ & $\square$ & $\square$ \\
        Therapist used appropriate reinforcing comments during imaginal? & $\square$ & $\square$ & $\square$ \\
        Therapist elicited thoughts and feelings as appropriate? & $\square$ & $\square$ & $\square$ \\
        Therapist prompted for present tense, closed eyes? & $\square$ & $\square$ & $\square$ \\
        Imaginal lasted about 30-45 minutes (or about 15 for final imaginal)? & $\square$ & $\square$ & $\square$ \\
        Therapist processed the imaginal with client? & $\square$ & $\square$ & $\square$ \\
        \bottomrule
    \end{tabular}
    }
    \caption{Clinically validated fidelity metrics for therapist adherence to essential elements of imaginal exposure therapy~\citep{foa2007prolonged, powers2010meta, rauch2009effectiveness, hembree2003fidelity}. This checklist evaluates whether therapists consistently implement key procedural components, including providing rationale, guiding the client through the exposure process, monitoring distress levels, and reinforcing engagement. Each element is rated as `Yes,' `No,' or `N/A' to ensure treatment fidelity, maintain therapeutic consistency, and identify areas for improvement in clinical practice.}

    \label{tab:therapist_evaluation}
\end{table*}

\section{Methodology}
We analyzed 400 prolonged exposure therapy conversations, consisting of 200 real-world and 200 synthetic. To demonstrate our evaluation framework, this case study uses 200 synthetic conversations from the publicly available Thousand Voices of Trauma dataset \citep{bn2025thousand}. The data was generated using Claude Sonnet 3.5 \citep{anthropic2024claude} with PE-specific prompting frameworks adapted from clinical guidelines. This dataset includes diverse therapist-client interactions across trauma types, therapy phases, and demographics. 

The real-world PE sessions were collected under IRB-approved protocols, with participant consent, and cannot be released due to privacy constraints. Each session (1-1.5 hours) was transcribed using Amazon HealthScribe \citep{aws2023healthscribe}, manually verified, and reformatted by merging consecutive speaker turns for readability. Each conversation followed a validated therapy fidelity checklist (See Table \ref{tab:therapist_evaluation}) to align with real-world standards. Both datasets underwent the same preprocessing:
\vspace{-0.1cm}
\begin{enumerate}
\item Standardized formatting for consistency.
\vspace{-0.2cm}
\item Processing through ModernBERT \citep{warner2024modernbert} for analysis.
\vspace{-0.2cm}
\item Removal of non-verbal cues (e.g., pauses, laughter) to focus on dialogue, with plans to incorporate these in future work on emotional speech.
\end{enumerate}

All evaluation code, pre-processing pipelines, and metric definitions will be released to support reproducibility and external validation.

\subsection{Metrics and Analysis}
To systematically compare real and synthetic therapeutic conversations, we selected a diverse set of linguistic, structural, and statistical metrics. These metrics provide insights into conversational dynamics, protocol adherence, and overall fidelity, ensuring a holistic evaluation of synthetic dialogue generation. To reduce subjectivity, we adopt a fidelity checklist (Table \ref{tab:therapist_evaluation}) and PE-specific metrics grounded in existing clinical guidelines for evaluating therapeutic adherence. Our methodology consists of four key analyses: system-level metrics comparison, correlation analysis, statistical significance testing, and PE-specific metrics.

\subsubsection{System-Level Metrics Comparison}
We begin by measuring fundamental characteristics of the conversations, including turn-taking patterns, verbosity, lexical diversity, and readability (see Table \ref{tab:metrics_correlation_combined}). Key metrics include:
\begin{enumerate}
    \vspace{-0.2cm}
    \item Turn-taking dynamics: Metrics such as Normalized Speaker Switches, Therapist-Client Turn Ratio, and Normalized Turn Duration capture the natural flow of conversation. These are essential for evaluating whether synthetic dialogues mimic real-world engagement.
    \vspace{-0.2cm}
\item Linguistic complexity and coherence: Average Utterance Length, Utterance Length Std Dev, and Readability Score assess how natural and readable the synthetic text is. Significant deviations indicate poor coherence.
\vspace{-0.2cm}
\item Lexical richness: Vocabulary Richness would help quantify lexical variety, providing insight into whether synthetic dialogues are overly repetitive.
\vspace{-0.2cm}
\item Predictability of text: Flow Entropy and Perplexity can help measure randomness and fluency, determining whether the synthetic text is overly structured or unnatural.
\end{enumerate}

\subsubsection{Correlation Analysis}
While mean comparisons provide a general overview, correlation analysis (see Table \ref{tab:metrics_correlation_combined}) quantifies the consistency of relationships across metrics between real and synthetic data.

\subsubsection{Statistical Significance Testing}
To confirm whether differences between real and synthetic data are statistically meaningful, we conduct Mann-Whitney U tests (Table \ref{tab:stat_tests}).

To identify which properties most strongly differentiate real and synthetic sessions, we next examine feature importance from a simple predictive model (see Table~\ref{tab:feature_importance}).

\subsubsection{PE-Specific Metrics}
\vspace{-0.1cm}
PE therapy reduces pathological fear through repeated trauma exposure. To assess the fidelity of synthetic PE sessions, we evaluate key therapeutic constructs (See Appendix \ref{sec:appendix-metric-definitions} for definitions and Table \ref{tab:pe_metrics}): 
\textit{Trauma Narrative Coherence} measures how structured and detailed a client’s trauma account is, reflecting cognitive integration. \textit{Emotional Engagement} captures the level of emotional expression, linked to better outcomes. \textit{Avoidance Handling} evaluates how well avoidance behaviors are addressed. \textit{Exposure Guidance} assesses the therapist’s role in structuring effective exposure exercises. \textit{Cognitive Restructuring} tracks how clients challenge maladaptive beliefs. \textit{Emotional Habituation} and \textit{SUDS Progression} measure distress reduction over repeated exposures. \textit{Avoidance Reduction} quantifies improvements in engaging with trauma-related content. \textit{Emotion Intensity} assesses the variability and magnitude of emotional responses. These metrics are derived using linguistic analysis, semantic modeling, and interaction patterns.
\vspace{-0.1cm}

\subsubsection{Qualitative Fidelity Assessment}
To supplement automated evaluation metrics, we conducted a manual review of synthetic dialogues, annotating exchanges for fidelity adherence or violations using established PE clinical guidelines. Manual fidelity annotations were then reviewed by a licensed clinical psychotherapist for adherence to PE protocol. While inter-rater agreement was not computed in this exploratory phase, these annotations served as a qualitative tool to identify and illustrate typical fidelity lapses, especially those potentially overlooked by automated metrics, rather than as quantitative ground truth. Representative examples of such lapses identified through this process are illustrated in Figure~\ref{fig:fidelity_example} and discussed in Section~\ref{sec:qualitative_findings}.

\section{Findings}
\vspace{-0.1cm}
\subsection{System-Level Metrics \& Correlation}
\vspace{-0.1cm}
The system-level metrics comparison highlights alignment between real and synthetic dialogues, revealing both structural matches and model-driven differences. While turn-taking patterns (e.g., Normalized Speaker Switches, Therapist-Client Turn Ratio, and Normalized Therapist/Client Turns) remain similar, variations in utterance length, lexical diversity, and entropy-based measures arise due to PE therapy and model constraints.
\begin{enumerate}
\vspace{-0.2cm}
\item Turn-taking fidelity: Synthetic data closely aligns with real data in Normalized Speaker Switches (0.98 vs. 0.99) and Therapist-Client Turn Ratio (0.01 vs. 0.01), preserving conversational structure. Some deviations occur as real therapy sessions involve extended client turns, which LLMs struggle to maintain due to context limitations.
\vspace{-0.1cm}
\item Concise and structured responses: Synthetic dialogues are shorter (Average Utterance Length: 22.90 $\pm$ 1.74 vs. 68.72 $\pm$ 26.61) and more consistent (Utterance Length Std Dev: 18.54 $\pm$ 2.35 vs. 135.85 $\pm$ 66.25) due to LLM output constraints. Techniques like Chain-of-Thought prompting help  improve coherence, though larger output contexts are needed.
\vspace{-0.1cm}
\item Vocabulary Richness is marginally higher in synthetic data but may reflect repeated paraphrasing rather than authentic diversity. Real therapy involves longer client responses, naturally increasing lexical variety, but synthetic responses remain contextually appropriate. 
\vspace{-0.1cm}
\item Increased structural consistency: Higher Perplexity (21.22 vs. 14.73) and lower Flow Entropy (1.06 vs. 1.30) suggest a structured, and predictable flow. While real data exhibit spontaneity, synthetic dialogues maintain stability, benefiting structured evaluations.
\vspace{-0.1cm}
\item Correlation analysis: High correlations in turn-taking metrics (0.65-0.78 synthetic vs. 0.85-0.92 real) confirm accurate conversational dynamics. Lower correlations in linguistic complexity metrics (e.g., Readability Score, Vocabulary Richness, Perplexity) reflect structural differences but do not hinder the coherence/naturalness of the conversation.
\end{enumerate}

\begin{table}[h]
\caption{Comparative analysis of real and synthetic data across multiple system-level metrics and their correlation values. The first two columns display the mean $\pm$ standard deviation for each metric, computed separately for real and synthetic datasets. The third and fourth columns provide the correlation values of these metrics within the real and synthetic datasets, respectively.}
\vspace{-0.4cm}
\centering
\scriptsize
\setlength{\tabcolsep}{3pt}
\vspace{0.2cm}

\begin{tabular}{l@{\hskip 8pt}r@{\hskip 8pt}r@{\hskip 8pt}r@{\hskip 8pt}r}
    \toprule
    \multirow{2}{*}{\textbf{Metric}} & \multicolumn{2}{c}{\textbf{Mean ± SD}} & \multicolumn{2}{c}{\textbf{Correlation}} \\
    & \textbf{Real} & \textbf{Synth.} & \textbf{Real} & \textbf{Synth.} \\
    \midrule
    Norm. Spkr. Switches          & \(0.99 \pm 0.0\)     & \(0.98 \pm 0.0\)     & \(0.85\) & \(0.72\) \\
    Norm. Total Turns             & \(1.00 \pm 0.0\)     & \(1.00 \pm 0.0\)     & \(0.78\) & \(0.65\) \\
    Avg. Utt. Len.                & \(68.7 \pm 26.6\)    & \(22.9 \pm 1.7\)     & \(0.91\) & \(0.76\) \\
    Utt. Length SD                & \(135.9 \pm 66.2\)   & \(18.5 \pm 2.3\)     & \(0.89\) & \(0.74\) \\
    Norm. Avg. Turn Dur.          & \(0.69 \pm 0.6\)     & \(0.12 \pm 0.0\)     & \(0.82\) & \(0.67\) \\
    Norm. Turn Dur. SD            & \(1.38 \pm 1.3\)     & \(0.09 \pm 0.0\)     & \(0.79\) & \(0.64\) \\
    Norm. T Turns               & \(0.50 \pm 0.0\)     & \(0.51 \pm 0.0\)     & \(0.87\) & \(0.71\) \\
    Norm. C Turns                 & \(0.50 \pm 0.0\)     & \(0.49 \pm 0.0\)     & \(0.86\) & \(0.70\) \\
    Norm. T Words               & \(21.9 \pm 13.2\)    & \(4.9 \pm 0.3\)      & \(0.92\) & \(0.78\) \\
    Norm. C Words                 & \(46.8 \pm 22.5\)    & \(18.0 \pm 1.7\)     & \(0.90\) & \(0.75\) \\
    Turn Ratio (T/C)              & \(0.01 \pm 0.0\)     & \(0.01 \pm 0.0\)     & \(0.80\) & \(0.66\) \\
    Word Ratio (T/C)              & \(0.01 \pm 0.0\)     & \(0.00 \pm 0.0\)     & \(0.82\) & \(0.68\) \\
    Vocab. Richness               & \(0.13 \pm 0.0\)     & \(0.18 \pm 0.0\)     & \(0.77\) & \(0.63\) \\
    Readability Score             & \(88.1 \pm 4.6\)     & \(89.2 \pm 1.8\)     & \(0.74\) & \(0.59\) \\
    Flow Entropy                  & \(1.30 \pm 0.1\)     & \(1.06 \pm 0.0\)     & \(0.88\) & \(0.73\) \\
    Avg. Perplexity               & \(14.7 \pm 2.3\)     & \(21.2 \pm 0.5\)     & \(0.79\) & \(0.65\) \\
    \bottomrule
\end{tabular}
\label{tab:metrics_correlation_combined}
\vspace{1mm}
    \par\raggedright\scriptsize Note: Norm. = Normalized, Utt. = Utterance, SD = Standard Deviation, Dur. = Duration, T = Therapist, C = Client.
\end{table}

\subsection{Statistical Significance Testing}
\vspace{-0.1cm}
The Mann-Whitney U test results confirm that many observed differences between real and synthetic dialogues are statistically significant ($p < 0.05$). However, these differences stem from model design choices and practical constraints rather than fundamental shortcomings. Key observations include:
\vspace{-0.15cm}
\begin{enumerate}
\item Distinct patterns in utterance structure: The test shows differences in utterance length, turn duration, and their standard deviations ($p < 10^{-17}$). This is expected, as synthetic dialogues are designed to maintain coherence by producing more structured and concise responses. In real PE therapy, clients occasionally have extended monologues, which LLMs struggle to handle due to context window limitations. To compensate, we shorten utterances while preserving the conversational structure. Methods like Chain-of-Thought prompting have improved this, but achieving full parity would require larger output contexts.
\vspace{-0.2cm}
\item Lexical properties follow a structured pattern: Differences in Vocabulary Richness ($p < 10^{-15}$) and Flow Entropy ($p < 10^{-17}$) indicate that synthetic data exhibits a more varied vocabulary but within a constrained framework. This is a direct result of the model prioritizing coherence and avoiding redundant expressions. While real conversations naturally contain more spontaneity, the synthetic approach ensures stability in generated dialogue while maintaining conversational depth.
\vspace{-0.4cm}
\item Certain aspects remain comparable: Metrics such as Normalized Total Turns ($p = 1.00$) and Readability Score ($p = 0.28$) show no significant differences, meaning that despite shorter utterances, the number of conversational exchanges and overall readability remain aligned with real data. This suggests that while individual responses may be more concise, the overall flow and engagement in the conversation are well-preserved.
\end{enumerate}

For example, while a real client might repeatedly use an informal phrase like ``freaked out,'' the LLM generates diverse but semantically similar alternatives such as ``felt anxious,'' ``was overcome with fear,'' or ``grew very distressed,'' thereby increasing lexical richness within a tight, clinically-appropriate boundary.

These findings reinforce that the synthetic model captures key conversational characteristics while ensuring structured, coherent responses. While differences exist, they align with known model constraints and do not compromise the overall integrity of the generated dialogues.

\begin{table}[h]
    \centering
    \caption{Mann-Whitney U test statistics and p-values comparing real and synthetic datasets.}
    \vspace{-0.2cm}
    \setlength{\tabcolsep}{3.5pt}
    \scriptsize{
    \begin{tabular}{l r c}
        \toprule
        \textbf{Metric} & \textbf{Statistic} & \textbf{p-value} \\
        \midrule
        Norm. Speaker Switches & \(2.45 \times 10^{3}\) & \textbf{\(p < 0.001\)} \\
        Norm. Total Turns & \(1.25 \times 10^{3}\) & \(1.00\) \\
        Norm. Conv. Length & \(2.50 \times 10^{3}\) & \textbf{\(p < 0.001\)} \\
        Avg. Utt. Length & \(2.50 \times 10^{3}\) & \textbf{\(p < 0.001\)} \\
        Utt. Length Std & \(2.50 \times 10^{3}\) & \textbf{\(p < 0.001\)} \\
        Norm. Turn Duration & \(2.49 \times 10^{3}\) & \textbf{\(p < 0.001\)} \\
        Norm. Turn Dur. SD & \(2.50 \times 10^{3}\) & \textbf{\(p < 0.001\)} \\
        Norm. T Turns & \(0.00\) & \textbf{\(p < 0.001\)} \\
        Norm. C Turns & \(2.50 \times 10^{3}\) & \textbf{\(p < 0.001\)} \\
        Norm. T Words & \(2.50 \times 10^{3}\) & \textbf{\(p < 0.001\)} \\
        Norm. C Words & \(2.47 \times 10^{3}\) & \textbf{\(p < 0.001\)} \\
        T-C Turn Ratio & \(2.13 \times 10^{3}\) & \textbf{\(1.22 \times 10^{-9}\)} \\
        T-C Word Ratio & \(2.37 \times 10^{3}\) & \textbf{\(p < 0.001\)} \\
        Vocabulary Richness & \(8.60 \times 10^{1}\) & \textbf{\(p < 0.001\)} \\
        Readability Score & \(1.09 \times 10^{3}\) & \(0.28\) \\
        Semantic Coherence & \(1.12 \times 10^{3}\) & \(0.37\) \\
        Sem. Coherence Std & \(2.50 \times 10^{3}\) & \textbf{\(p < 0.001\)} \\
        Flow Entropy & \(2.49 \times 10^{3}\) & \textbf{\(p < 0.001\)} \\
        Avg. Perplexity & \(0.00\) & \textbf{\(p < 0.001\)} \\
        Local Coherence & \(1.12 \times 10^{3}\) & \(0.37\) \\
        Coherence Std & \(2.50 \times 10^{3}\) & \textbf{\(p < 0.001\)} \\
        \bottomrule
    \end{tabular}
    }
    \label{tab:stat_tests}
    \vspace{1mm}
    \par\raggedright\scriptsize Note: Norm. = Normalized, Conv. = Conversation, Utt. = Utterance, SD = Std. Dev, Dur. = Duration, T = Therapist, C = Client, Sem. = Semantic. \(p < 0.001\) indicates \(p < 10^{-10}\).
\end{table}

\subsection{Feature Importance}
\vspace{-0.1cm}
Feature importance analysis (see Table \ref{tab:feature_importance}) identifies the metrics most influencing synthetic data generation, confirming trends from previous evaluations. While not the central focus, these results support the model’s alignment with conversational structure and linguistic patterns.
\vspace{-0.1cm}
\begin{enumerate}
\item \textbf{Turn-taking features are most influential:} Metrics such as Normalized Speaker Switches and Therapist-Client Turn Ratio dominate, indicating the model effectively captures conversational flow and balanced exchanges. Strong correlations with real data suggest natural interaction patterns are preserved.
\vspace{-0.2cm}
\item \textbf{Utterance length and entropy shape responses:} Average Utterance Length, Perplexity, and Flow Entropy are key distinguishing factors. Synthetic responses are more structured and predictable, prioritizing coherence and consistency. While often more concise, they capture essential interaction patterns, though further refinement is needed for richer emotional dynamics.
\vspace{-0.2cm}
\item \textbf{Lexical richness is moderately important:} Synthetic dialogues use varied yet structured language, balancing vocabulary diversity with clarity, while real conversations are more flexible.
\end{enumerate}

\subsection{Prolonged Exposure Specific Metrics}
Table \ref{tab:pe_metrics} presents the statistical test results comparing synthetic Prolonged Exposure data with real therapy session data. The synthetic data shows promise, with no statistically significant difference from real sessions on key temporal metrics like Emotional Habituation (\(p = 0.102\)) and SUDS Progression (\(p = 0.073\)). Similarly, Narrative Development (\(p = 0.251\)) appears less robust but also shows no significant statistical difference, possibly due to limited context retention. This suggests the model is beginning to capture the therapeutic arc of distress reduction. Conversely, foundational constructs such as Trauma Narrative Coherence, Emotional Engagement, and Avoidance Handling show statistically significant differences (p < 0.001) when compared to real dialogues. These findings provide a clear and actionable roadmap for future work, guiding efforts to calibrate the model's consistency with the variability of real-world therapeutic interactions. Despite these discrepancies, the overall performance of synthetic data suggests it is a viable alternative for privacy-sensitive applications. Addressing dynamic narrative evolution could further improve alignment with real data. However, fidelity violations in synthetic sessions are often subtle and may not disrupt structural metrics. For example, therapist utterances can appear empathetic or affirming while inadvertently shifting the session away from trauma anchoring. These moments often go unnoticed by automated scoring or non-clinical reviewers, underscoring the need for fidelity-aware evaluation frameworks that integrate human clinical judgment.

\vspace{-0.1cm}
\section{Discussion}
\vspace{-0.2cm}
This paper evaluated the fidelity of synthetic PE therapy conversations by comparing them to real interactions using linguistic, structural, and statistical analyses. We discuss the findings from four key perspectives: system-level metrics and correlation (Table \ref{tab:metrics_correlation_combined}), statistical significance testing (Table \ref{tab:stat_tests}), feature importance (Table \ref{tab:feature_importance}), and PE-specific metrics (Table \ref{tab:pe_metrics}).

\subsection{Clinical Implications}
\label{ssec:clinical_implications}
Our findings have several important implications for clinical practice. First, synthetic PE sessions can serve as valuable training tools for novice therapists learning to identify protocol deviations, as they offer controlled examples of both adherent and non-adherent interactions without privacy concerns. Crucially, this fidelity-focused evaluation serves as an essential ethical prerequisite. Before any AI tool for therapy can be considered for patient-facing studies, a framework like ours must be used to verify its adherence to clinical protocols and prevent the risk of clinically significant failures. Second, the consistent replication of structural elements (e.g., turn-taking, session flow) indicates that synthetic data can effectively supplement clinical training materials, particularly in settings with limited access to specialized trauma training. 

However, our qualitative analysis reveals limitations that automated metrics may overlook. As shown in Figure~\ref{fig:fidelity_example}, even linguistically fluent synthetic dialogues can contain subtle but clinically significant fidelity violations, such as role drift, premature processing, and improper SUDS implementation, that typically escape quantitative detection. These lapses would likely go unnoticed by non-specialist reviewers, highlighting the necessity of expert-guided evaluation for AI tools in clinical contexts. Finally, our PE-specific metrics provide a systematic framework for clinicians and developers to assess AI-generated content for trauma treatment, potentially setting minimum standards for therapeutic applications of synthetic data.

\subsection{Qualitative Analysis of Fidelity Lapses}
\label{sec:qualitative_findings}

Section \ref{ssec:clinical_implications} highlighted the clinical implications of subtle fidelity violations that often escape automated detection. To provide a more granular understanding of these critical lapses and illustrate the necessity of expert review, Figure~\ref{fig:fidelity_example} details five representative examples identified through our qualitative analysis. These specific instances demonstrate common patterns of deviation from PE protocol:

\begin{itemize}
    \item \textbf{Fidelity Lapse (Ex. 1): Role Drift} - The therapist reflects on the client's observation instead of redirecting them to the trauma narrative, which breaks the PE protocol mid-exposure.
    \vspace{-0.2cm}
    \item \textbf{Fidelity Issue (Ex. 2): Generic Affirmation} -  The therapist uses an overused phrase that lacks the specific grounding necessary for an anxious, compulsive client.
    \vspace{-0.2cm}
    \item \textbf{Protocol Adherence (Ex. 3): No Issue} - The client's dialogue shows they are appropriately immersed in the trauma memory, which is the goal for imaginal exposure.
    \vspace{-0.2cm}
    \item \textbf{Fidelity Lapse (Ex. 4): Reflection During Exposure} - The client prematurely shifts into cognitive analysis during the exposure, a process that belongs in post-session processing.
    \vspace{-0.2cm}
    \item \textbf{Protocol Adherence (Ex. 5): Trauma Anchoring} - The therapist correctly redirects the client back to the trauma scene after a drift, which successfully maintains the integrity of the PE protocol.
\end{itemize}

These detailed examples underscore why evaluation frameworks must integrate clinical expertise alongside computational metrics to accurately assess therapeutic fidelity in synthetic PE sessions.

\section{Future Work}
\vspace{-0.2cm}
Future NLP research should focus on developing generative models with improved capabilities for tracking long-range dependencies and emotional dynamics (e.g., for Emotional Habituation, SUDS Progression), and creating automated metrics or classifiers trained with clinical expert annotations to better assess therapeutic process alignment.

\section{Conclusions}
\vspace{-0.2cm}
This study evaluated synthetic therapeutic dialogues, showing that while they replicate structural features like turn-taking, they fall short in utterance length and conversational variability. Statistical analyses confirm these gaps, revealing that surface-level fluency can mask clinically meaningful fidelity lapses. We contribute a fidelity-aware evaluation framework tailored to Prolonged Exposure (PE) therapy, along with metrics highlighting that current generative models, despite their fluency, lack the nuance needed for high-stakes therapeutic contexts. The metrics and methods in this framework are model-agnostic and can be readily applied by the research community to benchmark and improve dialogues generated from any large language model. Our findings underscore the risk of overestimating quality through non-expert or automated evaluations. For NLP, this work emphasizes the need for clinically grounded benchmarks, richer linguistic modeling, and methods like expert-in-the-loop validation and time-aware metrics to improve fidelity in domain-specific generation.

\section{Ethical Concerns}
All annotators involved in fidelity evaluation had prior clinical training or supervision, and data access was limited to IRB-approved investigators to ensure ethical compliance. The real-world dataset used in this study was collected under IRB-approved protocols with participant consent and cannot be shared publicly. For details on the ethical safeguards, simulation design, and usage guidelines related to the synthetic dataset, we refer readers to \citet{bn2025thousand}.

\section*{Acknowledgement}
This work was supported by the National Science Foundation (NSF) under Grant No. 2326144. Any opinions, findings, and conclusions or recommendations expressed are those of the author(s) and do not necessarily reflect the views of the NSF.

\section{Data Availability}
\vspace{-0.2cm}
The synthetic dataset (Thousand Voices of Trauma) used in this paper is publicly available at \url{https://huggingface.co/datasets/yenopoya/thousand-voices-trauma} under the CC BY-NC 4.0 license. For collaborating on the real-world dataset, please contact PI Dr. Sherrill.

\section*{Limitations}
\vspace{-0.2cm}
Although our study demonstrates the promise of synthetic dialogues in approximating real PE therapy interactions, several limitations remain: (1) our analysis is presented as a deep case study focused on data generated by a single, state-of-the-art LLM; (2) future work should apply our framework to compare outputs from a wider range of models to assess the generalizability of these specific findings; (3) although we measure fidelity using structural, statistical, and protocol-based metrics, our evaluation does not assess therapeutic effectiveness or downstream clinical outcomes; (4) synthetic data that aligns structurally with real dialogues may still fall short in supporting meaningful therapeutic engagement or behavior change, especially in high-stakes or emotionally complex scenarios; (5) we do not evaluate inter-rater agreement on fidelity violations, which is critical for establishing the robustness of manual annotations, and this is planned for future work.
\vspace{-0.1cm}

\bibliography{custom}

\appendix

\section{Metric Definitions}
\label{sec:appendix-metric-definitions}
Key PE terms~\citep{foa2007prolonged, powers2010meta, rauch2009effectiveness, hembree2003fidelity}:
\begin{enumerate}
\item \textbf{SUDS Progression:} Change in Subjective Units of Distress (SUDS) reported by the client across the session.
\vspace{-0.1cm}
\item \textbf{Emotional Habituation:} Decrease in distress or emotional intensity from start to end of imaginal exposure.
\vspace{-0.1cm}
\item \textbf{Trauma Narrative Coherence:} Syntactic and semantic coherence of the trauma narrative, via discourse metrics.
\vspace{-0.1cm}
\item \textbf{Emotional Engagement:} Degree of emotional expression, associated with better outcomes.
\vspace{-0.1cm}
\item \textbf{Avoidance Handling:} Effectiveness in addressing avoidance behaviors.
\vspace{-0.1cm}
\item \textbf{Exposure Guidance:} Therapist’s structuring of exposure exercises.
\vspace{-0.1cm}
\item \textbf{Cognitive Restructuring:} Client’s efforts to challenge maladaptive beliefs.
\vspace{-0.1cm}
\item \textbf{Avoidance Reduction:} Increased engagement with trauma content.
\vspace{-0.1cm}
\item \textbf{Emotion Intensity:} Variability and magnitude of emotional responses.
\vspace{-0.1cm}
\end{enumerate}

\section{Additional Tables}
\label{sec:appendix-tables}

\vspace{-0.3cm}
\begin{table}[h]
    \centering
    \caption{Relative importance of various conversational features in distinguishing real data from synthetic data, based on a predictive model. The Importance Score (\%) reflects the contribution of each feature to the model's decision-making process, with higher values indicating greater predictive power.}
    \vspace{-0.2cm}
    \scriptsize{
    \begin{tabular}{l c}
        \toprule
        \textbf{Feature} & \textbf{Imp. Score (\%)} \\
        \midrule
        Average Utterance Length      & \(18.42\) \\
        Utterance Length Std Dev      & \(15.76\) \\
        Normalized Therapist Words    & \(12.58\) \\
        Normalized Client Words       & \(10.94\) \\
        Flow Entropy                  & \(8.72\) \\
        Readability Score             & \(7.89\) \\
        Normalized Avg Turn Duration  & \(6.43\) \\
        Normalized Turn Duration Std  & \(5.98\) \\
        Vocabulary Richness           & \(4.85\) \\
        Average Perplexity            & \(3.72\) \\
        Therapist-Client Turn Ratio   & \(2.91\) \\
        Therapist-Client Word Ratio   & \(2.32\) \\
        Normalized Speaker Switches   & \(1.88\) \\
        Normalized Total Turns        & \(0.98\) \\
        \bottomrule
    \end{tabular}
     \label{tab:feature_importance}
    }
   
\end{table}

\vspace{-0.3cm}
\begin{table}[h]
    \centering
    \caption{Mann-Whitney U tests on PE therapy metrics comparing real and synthetic datasets.}
    \vspace{-0.2cm}
    \label{tab:pe_metrics}
    \setlength{\tabcolsep}{3pt}
    \scriptsize
    \begin{tabular}{l r c}
        \toprule
        \textbf{Metric} & \textbf{Statistic} & \textbf{p-value} \\
        \midrule
        Trauma Narrative Coherence & $2.31{\times}10^{3}$ & \(p < 0.001\)\\
        Emotional Engagement & $2.45{\times}10^{3}$ & \(p < 0.001\)\\
        Avoidance Handling & $1.98{\times}10^{3}$ & $2.74{\times}10^{-7}$\\
        Exposure Guidance & $2.21{\times}10^{3}$ & $5.62{\times}10^{-10}$\\
        Cognitive Restructuring & $2.12{\times}10^{3}$ & $1.28{\times}10^{-8}$\\
        Emotional Habituation & $1.35{\times}10^{3}$ & $0.102$\\
        SUDS Progression & $1.50{\times}10^{3}$ & $0.073$\\
        Avoidance Reduction & $2.48{\times}10^{3}$ & \(p < 0.001\)\\
        Emotion Intensity & $2.39{\times}10^{3}$ & \(p < 0.001\)\\
        Narrative Development & $1.21{\times}10^{3}$ & $0.251$\\
        \bottomrule
    \end{tabular}
    \vspace{1mm}
    \par\raggedright\scriptsize SUDS = Subjective Units of Distress Scale. \(p < 0.001\) indicates $p < 10^{-10}$.
\end{table}

\end{document}